\documentclass[conference]{IEEEtran}
\usepackage{cite}
\usepackage{amsmath,amssymb,amsfonts}
\usepackage{subcaption}
\usepackage{hyperref}
\usepackage{float}
\usepackage{graphicx}
\usepackage[ruled,vlined, linesnumbered]{algorithm2e}
\usepackage{textcomp}
\usepackage{xcolor}
\usepackage{xurl}
\def\BibTeX{{\rm B\kern-.05em{\sc i\kern-.025em b}\kern-.08em
    T\kern-.1667em\lower.7ex\hbox{E}\kern-.125emX}}
\begin{document}

\title{Sky Computing: Accelerating Geo-distributed Computing in Federated Learning}

\author{\IEEEauthorblockN{Jie Zhu\textsuperscript{*}}
\IEEEauthorblockA{\textit{HPC-AI Technology Inc.} \\
Equal contribution\textsuperscript{*} \\
contact@hpcaitech.com}

\and

\IEEEauthorblockN{Shengui Li\textsuperscript{*}}
\IEEEauthorblockA{\textit{HPC-AI Technology Inc.} \\
Equal contribution\textsuperscript{*} \\
lisg@hpcaitech.com}

\and

\IEEEauthorblockN{Yang You}
\IEEEauthorblockA{\textit{National University of Singapore} \\
Work done at HPC-AI Technology Inc.\\
youy@comp.nus.edu.sg}
}

\maketitle

\begin{abstract} 
Federated learning is proposed by Google to safeguard data privacy through training models locally on users' devices. However, with deep learning models growing in size to achieve better results, it becomes increasingly difficult to accommodate the whole model on one single device. Thus, model parallelism is then used to divide the model weights among several devices. With this logic, the approach currently used evenly allocates weights among devices. However, in reality, a computation bottleneck may occur resulting from variant computing power of different users’ devices. To address this problem, load balancing is needed to allocate the model weights based on the computational capability of the device. In this paper, we proposed Sky Computing, a load-balanced model parallelism framework to adaptively allocate the weights to devices. Sky Computing outperforms the baseline method by 55\% in training time when training 160-layer BERT with 64 nodes.
The source code can be found at  \href{https://github.com/hpcaitech/SkyComputing}{\underline{https://github.com/hpcaitech/SkyComputing}}.

\end{abstract}

\begin{IEEEkeywords}
Federated Learning, Model Parallelism, Load Balance
\end{IEEEkeywords}

\section{Introduction} 

With the bloom of deep learning, the sizes of the models keep increasing. This is especially significant in Natural Language Processing area as models like BERT \cite{devlin-etal-2019-bert} and GPT-3 \cite{NEURIPS2020_1457c0d6} emerged with billions of parameters. These models have shown their capabilities of exploring and processing the vast amount of semantics embedded in the text, and delivering the state-of-the-art performance with which small models cannot compete. Even though some methods such as knowledge distillation \cite{44873} and model compression \cite{8953493} have been used to speed up the training, it is undeniable that these methods generally cause a decrease in performance. Meanwhile, we expect that more powerful computational capability will be brought by new hardware and more large-scale datasets will come out in the future. Thus, the increasing trend of model size is expected to continue.

At the same time, with the increasing scalability of applications and the explosive growth of computational capability of hardware, more and more companys 
tend to combine the local devices or traditional cloud resources. This geo-distributed computing will be extremely significant in deep learning, as we mentioned above, the size of model is continuously expanding and thus need more hardware resources. 

In general, training these models requires a large amount of data which are collected by many companies from their users. For example, the text input applications can train a prediction model which recommends the next word for sentence completion. They can use the current sentence as the input data and the user-selected words as the label to train the model. In this way, the model can better capture user behavior and does not require extra data from external parties. 

However, as concerns over data privacy are raised worldwide, it is difficult to directly collect and store the user data in the servers without the user's consent. To solve this problem, federated learning was proposed by Google which introduces a mechanism for local model training on user's devices without sharing the data with the service providers \cite{McMahan2017CommunicationEfficientLO}. This ensures that data is only kept on the user's device and stays intact. One disadvantage of federated learning is that the users' devices are generally mobile phones, tablets, and personal computers, and model training is limited by the device hardware specifications, especially CPU, GPU, and RAM. It is thus challenging to train a large-scale model on these devices since it is impossible to fit the entire model into one single device and there may exist computation bottlenecks due to the variance in computational capability. 

Since the heterogeneity of training equipment is relatively high, we shall consider it as an perfect situation for geo-distributed computing. And thus we can reduce the training time by accelerating geo-distributed computing. 

In this paper, we proposed a new federated learning framework Sky Computing which exploits the feature of geo-distributed computing by using a load-balanced strategy in model parallelism to train large-scale models on the users' devices in federated learning. Model parallelism distributes the weights on the devices. With more devices participating in federated learning, the average size of model weights allocated to each device is smaller, making it possible to train large-scale models. Load balancing is an efficient method in geo-distributed computing, and it is necessary for model-parallel training as the relatively slow devices can slow down the entire training process and incur the computation bottleneck. 

It is not ideal if too much computation is allocated to devices with weaker computing power. In federated learning, the users' devices participating in model training are also changing dynamically and their computing power is unpredictable. Thus, we cannot pre-allocate the model weights to these devices. In contrast, our Sky Computing can adaptively allocate the model layers to the devices based on the model information and device performance. In this way, the Sky Computing can eliminate the computation bottleneck and reduce the training time. In our experiments to train a 160-layer BERT, our approach can outperform the baseline approach by 55\% in terms of training time when using 64 nodes. The source code can be found at \href{https://github.com/hpcaitech/SkyComputing}{\underline{https://github.com/hpcaitech/SkyComputing}}.

\section{Background}

\subsection{Geo-distributed Computing}

The growing ubiquity of computing devices, including smart phones and cloud servers produced large and fluctuating volumes of data and therefore required high-speed, highly available and resource efficient data processing to ensure low response times for specific activities. Plus, aggregating and processing data at a centralized cloud platform is not adequate to meet the requirements of many problems, especially considering data privacy. And the geo-distributed computing, which connects devices at different levels together, is a perfect solution to these two problems. The heterogeneity of computing resources becomes the major hinder to design algorithm and allocation work load, but in the other hand it also could be exploited as a feature. 
    
\subsection{Federated Learning}

Federated learning is a paradigm of collaborative learning with distributed training data \cite{McMahan2017CommunicationEfficientLO}. In standard machine learning, the training data is stored in a centralized location and the data is sent to one or more machines for training. However, in commercial scenarios, engineers need to collect data from the users before the model can be trained. Thus, the standard mechanism cannot be applied with increasing concerns over data privacy.

\begin{figure}
    \centering
    \includegraphics[width=0.9\columnwidth]{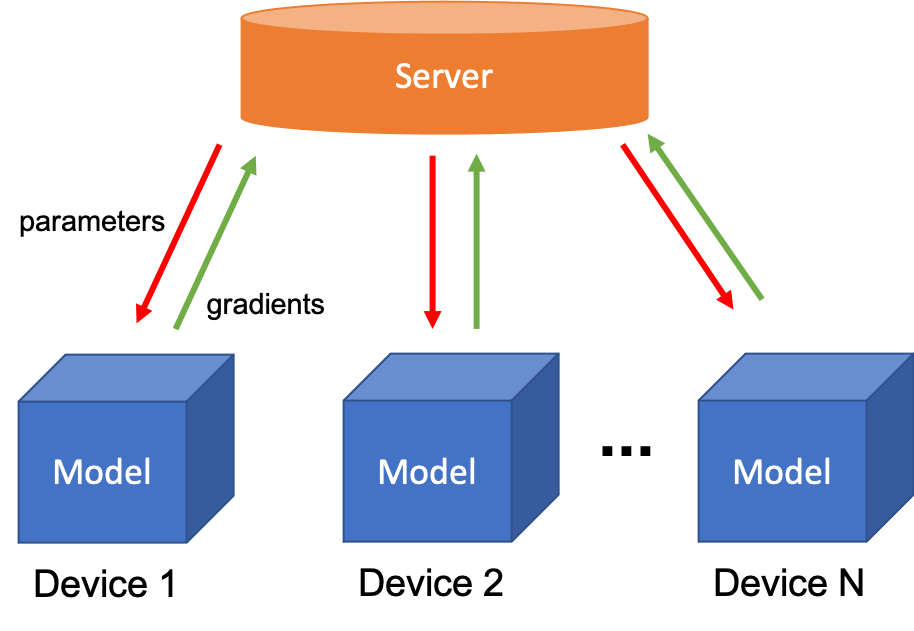}
    \caption{Architecture for Federated Learning}
    \label{fig:arch-fed-learning}
\end{figure}

Federated learning was proposed to replace the centralized training fashion with a decentralized training mechanism. The gist of federated learning is that the users can train the model locally on their devices without communicating personal data with others. In current methods, each user's device holds a copy of the model weights and uses the local data to train the model. To perform a global update of the model, only the gradients are passed back to the central server using encrypted communication. A back-propagation algorithm is performed to update the weights on the remote server using the Federated Averaging algorithm \cite{McMahan2017CommunicationEfficientLO}. The updated model weights are then broadcast back to the users' devices to update the local model as shown in Figure \ref{fig:arch-fed-learning}. In this way, the devices can collaboratively learn a shared and smarter prediction model while the users' data are kept invisible from the external parties to safeguard user privacy.

\subsection{BERT}

BERT stands for Bidirectional Encoder Representations from Transformers and is one of the state-of-the-art deep learning models for Natural Language Processing \cite{devlin-etal-2019-bert}. BERT yields superior performance in language tasks such as text classification, translation, and text synthesis and has been widely transferred to other fields such as Computer Vision. Its core module is the encoder layer, which relies on the self-attention mechanism to learn text representation. Self-attention is intrinsical to calculate the output tensor as a weighted sum of the input tensors. The encoder layer provides the self-attention mechanism to explore the correlation between words in a sentence and to better capture the text semantics in different contexts. The encoder layers can be succeeded by various projection heads for different downstream tasks.

In the encoder layer, the self-attention layer plays a pivotal role in understanding the context embedded in the sentence. The input data of the self-attention layer consists of a batch of sentences in the shape of $(b, n, h)$ where $b$ is the batch size, $n$ is the number of tokens in the sentence and $h$ is the hidden size of the tensor for each token. This embedding tensor is used to compute other three tensors: query $Q$, key $K$, and value $V$ by multiplying with three different matrices of the same shape $(h, v)$ where $h$ is the token hidden size and $v$ is the size of the attention heads. The output of the self-attention layer is obtained as shown in the equation \ref{eq: self-attention} where $d_{k}$ is the dimension of the key \cite{devlin-etal-2019-bert}. This mechanism allows each token to reference the semantic embedding of other tokens in the same sentence and calculate their relative importance in the context.

\begin{equation}
\text{Attention}(Q, K, V) = \text{softmax}(QK^{T}/\sqrt{d_k})V
\label{eq: self-attention}
\end{equation}

\begin{figure}
    \centering
    \includegraphics[width=0.9\columnwidth]{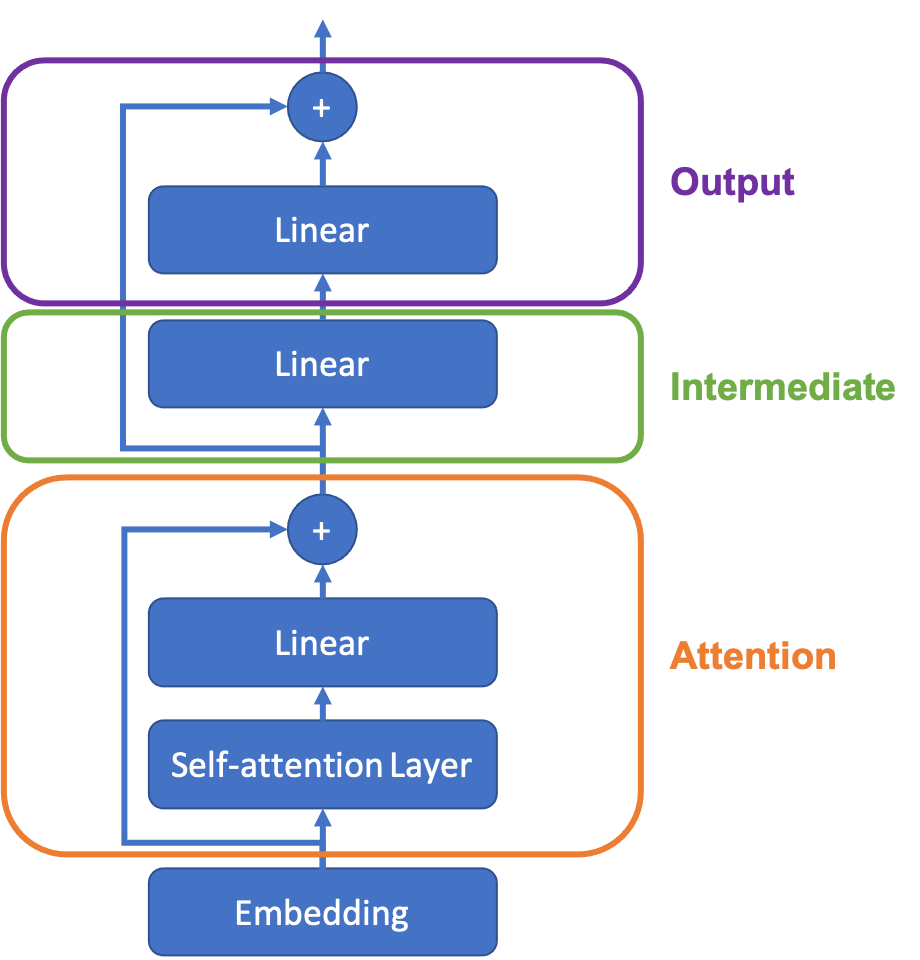}
    \caption{Architecture for BERT Encoder Layer}
    \label{fig:arch-bert-encoder-layer}
\end{figure}

Besides the self-attention layer, there are three more linear layers with residual connection in the encoder layer. The first linear layer keeps the dimension of the output tensor $h$ unchanged while the second linear layer projects the output tensor to a higher dimension $i$ and the third linear layer projects it back to the hidden size $h$. The architecture of the encoder layer can thus be divided into three components: attention, intermediate, and output as shown in Figure \ref{fig:arch-bert-encoder-layer}. The linear layer projection and the residual connection can embed the learned semantics into the output tensor and make the whole model deeper \cite{dong2021attention}.

Even though BERT can produce impressive results in many tasks, the time complexity of the self-attention layer is $O(n^2)$ where $n$ is the sequence length and the linear layers have high hidden dimension, posing challenges to the training speed \cite{devlin-etal-2019-bert}. Moreover, BERT has a large model size. A 24-layer BERT-Large model has 345 million parameters, making it difficult to train BERT on a single GPU.

\subsection{Model Parallelism}
To speed up the model training, parallel computing is integrated to increase the training throughput and reduce training time. There are mainly two paradigms of parallelism in machine learning, namely data parallelism and model parallelism \cite{10.1145/3320060}. 

Data parallelism is the mainstream method to train a model. In this paradigm, data is split and distributed to the workers to perform parallel computing. Each worker (GPU) holds a full copy of the model and trains on its data. It performs local backward-propagation to calculate the gradients of parameters. All-reduce is then applied to the gradients on all workers to perform weight updates. In this manner, we can increase the data throughput by having more workers to speed up the training.

\begin{figure}
    \centering
    \includegraphics[width=0.9\columnwidth]{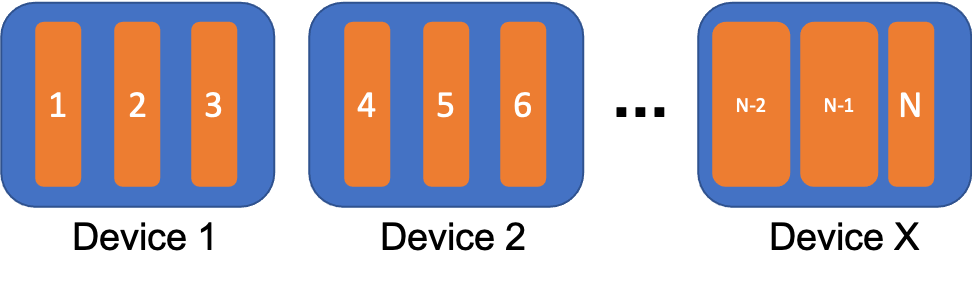}`
    \caption{Model Parallelism by Layer}
    \label{fig:model-parallel-by-layer}
\end{figure}

However, this method does not work if the size of the model goes beyond the memory limit of a single worker. To address this problem, model parallelism should be used instead, in which the model is split into several parts. There are mainly two paradigms of model parallelism, namely by layer and by tensor. Model parallelism by layer means that the model is split layerwise as shown in Figure \ref{fig:model-parallel-by-layer} when the memory of a single device is insufficient. This is often used together with pipeline parallelism and some examples include GPipe \cite{NEURIPS2019_093f65e0} and PipeDream \cite{10.1145/3341301.3359646}.  Meanwhile, model parallelism by tensor is to split the weight tensor among the devices. One example of this paradigm is Megatron \cite{shoeybi2020megatronlm} which uses tensor splitting on the embedding layer and the encoder layer of Transformer \cite{NEURIPS2019_9d63484a} to scale up training. 

\section{Model Parallelism in Federated Learning}
\begin{figure} 
    \centering
    \includegraphics[width=0.9\columnwidth]{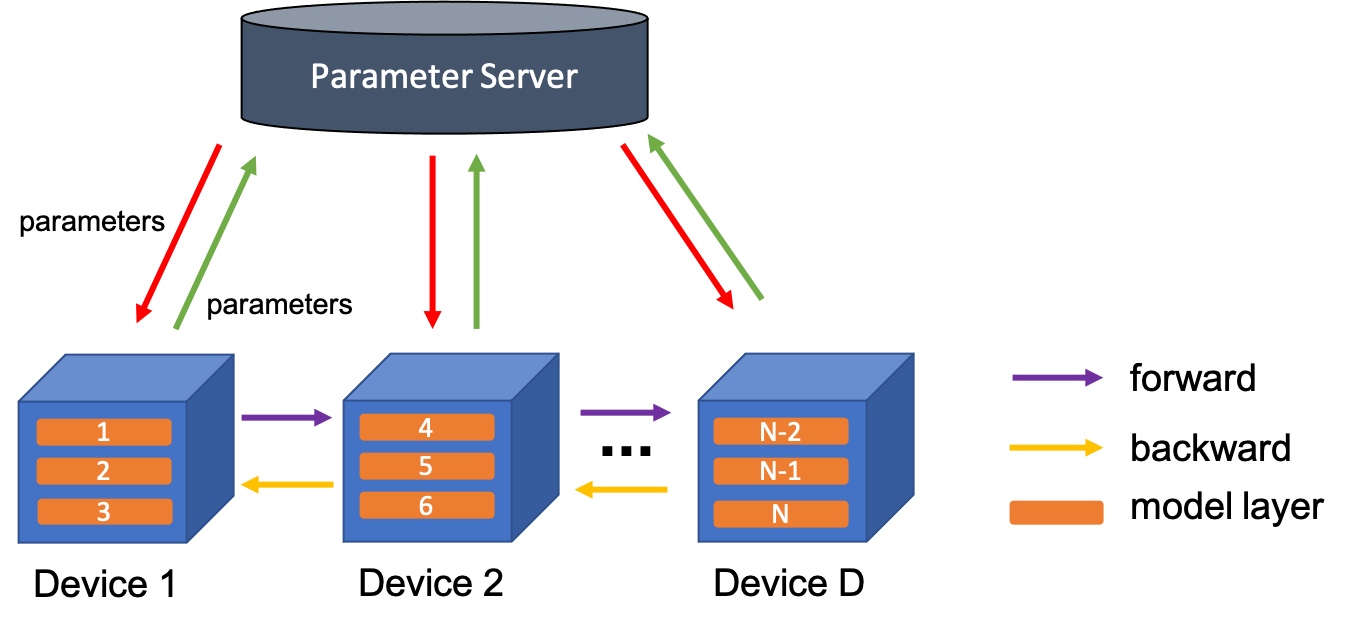}`
    \caption{Model Parallelism in Federated Learning}
    \label{fig:model-parallelism-in-federated-learning}
\end{figure}

In current federated learning methods \cite{McMahan2017CommunicationEfficientLO}, \cite{ 47976}, \cite{10.1145/3298981}, each device needs to accommodate the whole model and the maximum model size is limited by the capability of a single device, hindering training of large-scale models. To tackle this problem, we proposed the idea of integrating model parallelism into federated learning to train large-scale models. As the number of devices involved in federated training can be dynamically changing, we adopted model parallelism by layer to allocate the model layers on different devices as shown in Figure \ref{fig:model-parallelism-in-federated-learning}. Similar to the normal federated learning, we also need a central server to host a full copy of the model weights as a parameter server \cite{NIPS2014_1ff1de77}. Once the layer allocation is determined, the device can fetch the weights of the allocated layers from the server. Each device computes its output tensor given an input from the previous device and the output tensor is passed to the next device. Once the last device has finished the forward pass, it starts to run the back-propagation algorithm. The gradients are calculated and the weights are updated in the reverse order. In this setting, the workers do not pass the gradients back to the server as the gradients are used to update the weights locally. The central server only communicates the model parameters with the devices as it distributes the model layers at the start of training and collects back the updated parameters at the end of training.

When we add more devices to this network, communication cost increases due to the passing of tensors between devices. However, the average computation on each device will decrease given that the total model size is constant. This is important in federated learning, especially in mobile application scenarios. Since the training is conducted on users' devices, it will undermine the overall user experience of their devices if training consumes too many hardware resources, in turn drawing dissatisfaction from the users.

\section{Geo-distributed Computing Optimization} 

As the devices have the different computing power and communication latency, it is of paramount importance to choose an allocation strategy to distribute the layers in federated learning. In this section, we present our Sky Computing to optimize the model parallelism in federated learning by using load-balanced strategy.

\subsection{Naive Approach} 
The simplest method to allocate the layers is to distribute an equal number of layers among the devices. This method is easy to implement but fails to take into account the computational power and communication latency of the devices. As a result, devices with weaker computing power and higher communication delays can cause a huge bottleneck in training. In addition, this method is not aware of the amount of memory required for training. The out-of-memory problem can occur if too many layers are allocated to a device with limited RAM. Therefore, more information about the device and the model is needed to allocate the layers more wisely.

\subsection{Benchmarking}

To balance the load of each node and avoid causing the out-of-memory problem, a benchmark stage is needed to test the performance of the devices, that is, to test how much load each device can bear. The benchmark stage is designed to gather information about the devices as well as the training model. We implemented two separate benchmark tests to extract the information about model structure and device capability. The collected information will be used to compute the allocation.

As the devices have different hardware configurations, it is necessary to understand their performance on the same benchmark. We need to know their relative latency to remove the bottleneck in training, the amount of available memory to avoid the out-of-memory problem. The latency of the devices is mainly caused by two aspects. The first is the delay of communication, which will increase the transfer time between devices. The second is the computational power of the devices. To get this information, we can send a request from the central server to each device and record the time interval between sending and receiving. Besides, the request will run a simple benchmark test to measure the time taken on each device. The benchmark test is to simply run the forward pass of a convolutional neural network or the first few layers of the training model for tens of iterations. This ensures that all devices have the same amount of computation for the sake of fairness. The ratio of time taken on these devices will be used as a reference of their relative computation power. To get the available memory on each device, the request will query for the system hardware information on the receiver device.

The model information is also essential in layer distribution. Similar to the device information, we also need to know how fast a layer can be computed and its memory usage. This can be done on the central server without interaction with the devices. For the first one, we can measure the number of floating-point operations of a layer. As this indicates the amount of computation in the forward pass, it can help us match the faster device with more computationally heavy layers and vice versa. For memory usage, we need to estimate the peak memory required for both forward pass and backward pass to avoid the out-of-memory problem. The memory consumption during training involves the input tensor, layer weights, gradients, output tensors, and other optimizer-associated states. The memory consumption of each component can be estimated by counting the number of floating numbers individually.

\begin{figure}
    \centering
    \includegraphics[width=0.9\columnwidth]{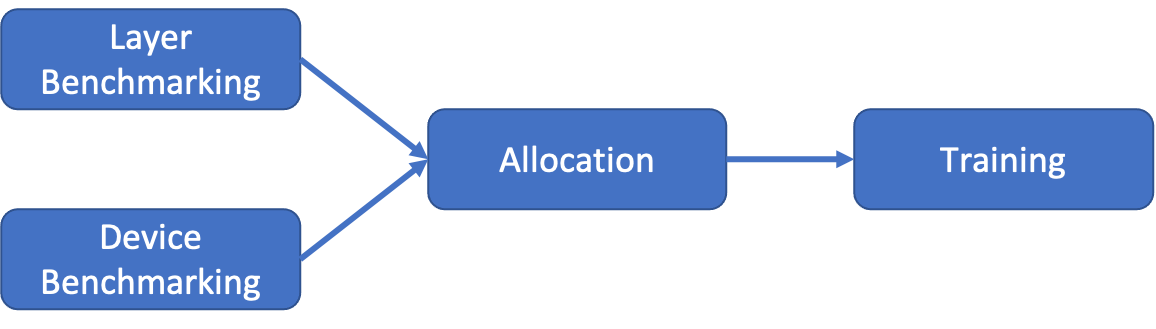}
    \caption{Training process in the Sky Computing}
    \label{fig:training-process}
\end{figure}

\subsection{Mathematical Model}
After getting the relevant hardware information above, we can use this information to write the allocation problem as a linear programming mathematical model. Suppose there are $L$ layers and $D$ devices. Use $x_{ij}$ as indicator variables such that $x_{ij} = 1$ iff $i$-th device contains $j$-th layer. The time for the $i$-th device to complete the benchmark can be denoted as $dt_i$. The available memory of $i$-th device is $dm_i$, the communication latency is $ct_i$. $lf_j$ and $lm_j$ represent the number of flops and size of the $j$-th layer, respectively. So the total workload of $i$-th device can be denoted as $w_{i} = dt_{i}\sum_{j=1}^{L}x_{ij}lf_{j} + ct_i$. Use $y_i$ represents the maximum index of the layers contained by $i$-th device, and $z_i$ represents the minimum. The complete model can be shown as:

\begin{alignat*}{2}
\min q &\quad &{} \tag{obj.}\\
\mbox{s.t.} \quad &\sum_{j=1}^{L}x_{ij}lm_{j} \le dm_{i}, &\quad &\forall i \in D\\
& y_i \ge jx_{ij}, &\quad &\forall i \in D, \forall j \in L \\
& z_i \le jx_{ij} + M(1-x_{ij}), &\quad &\forall i \in D, \forall j \in L \\
& y_{i} - z_{i} = \sum_{j=1}^{L} x_{ij} - 1, &\quad &\forall i \in D \\
& \sum_{i=1}^{D} x_{ij} = 1, &\quad &\forall j \in L \\ 
& q \ge dt_{i}\sum_{j=1}^{L}x_{ij}lf_{j} + ct_i, &\quad &\forall i \in D \\
& x\text{ is binary}, y, z\text{ are non-zero integer}
\end{alignat*}

Since it is a Mixed Integer Programming problem, we cannot get an optimal solution in polynomial time. With the increasing number of model layers as well as devices, the cost of obtaining the optimal solution is unacceptable. Therefore, we proposed a heuristic allocation approach.

\subsection{Heuristic Allocation}

For NP-hard problems, we can construct a heuristic algorithm to quickly obtain a feasible solution within an acceptable gap. In this paper, we designed a heuristic allocation algorithm that is only applicable in this problem, and we found in the experiment that the gap between the heuristic allocation and the optimal allocation is merely 10\% in the worst case. 

Our goal is to allocate layers such that each device has a similar workload. To achieve this goal, there are two conditions. Firstly, the layers allocated do not occupy more memory than the device memory limit. This is to make sure that training can at least be run even though the load is not balanced. Secondly, the workload on the devices should be similar to avoid computation bottlenecks and speed up training. Thus, we split the allocation into two sub-stages which are coarse allocation and workload fine-tuning.

From the previous benchmarking stage, we have gathered information about the devices and the model. Suppose we have $L$ layers and $D$ devices in the federated learning network, we let $dt$, $dm$ and $ct$ be arrays of length $D$ where $dt$ stores the time ratio for the benchmarking test, $dm$ stores the memory available on the devices and $ct$ stores the communication time of devices. The $i$-th element in these arrays refers to the time result or memory of the $i$-th device. $lf$ and $lm$ are arrays of length $L$ which stores the number of FLOPs and required memory of each layer.

In coarse allocation stage as shown in Algorithm \ref{algo:coarse-allocation}, we firstly initialize a partition index $pi$. The partition index $pi$ is an array of integers of length $D+1$. The element in the partition index refers to the index of the layer in the model. Thus, the partition index specifies the range of layers allocated to each device as the $i$-th device will be allocated with \texttt{pi[i]} to \texttt{pi[i+1]-1} layers. The partition index is initialized such that each device has the same number of layers. The layer allocation is then turned into a partitioning problem. We need to change the number of layers on each device to fulfill the memory requirement such that all devices have sufficient memory for the layers allocated. The strategy in the coarse allocation stage is that a device should take in more layers if it has enough memory and gives away some layers when the memory limit is exceeded. At the same time, it needs to ensure that each device has at least one layer. Our algorithm will iteratively update the partition index until the memory requirements are fulfilled on all devices. At this stage, we should be able to ensure training can be run successfully without concerns about out-of-memory problems.

\begin{algorithm}
\SetAlgoLined
\KwResult{pi }
// dt: device time, dm: device memory \\
dt, dm, ct = device\_benchmark()\; 

// lf: layer flops, lm: layer memory \\
lf, lm = model\_benchmark()\;

// pi: partition index \\
pi = initialize\_partition\_index()\;

// am: allocated memory on devices \\
am = compute\_allocated\_memory(lm, pi)\; 
\texttt{\\}
    
\While{true}{
    \uIf{is\_memory\_satisfied}{
        return pi\;
        }
    \texttt{\\}
    
    \For{$i \gets 0$  \KwTo $D$-2} {
        \While{am[$i$] $>$ dm[$i$] \&\& get\_num\_layer(pi, $i$) $>$ 1 }{
            pi[$i$+1] -= 1\;
            update\_allocated\_memory()\;
        }
        
        \While {dm[$i$] $>$ am[$i$] + lm[pi[$i$+1]] \&\& get\_num\_layer(pi, $i$+1) $>$ 1}{
            pi[$i$+1] += 1\;
            update\_allocated\_memory()\;
        }
    }
    \texttt{\\}
    
    \uIf{is\_pi\_unchanged}{
        raise exception(Cannot fulfill memory requirement)\;
    }
}
\caption{coarse allocation by memory}
\label{algo:coarse-allocation}
\end{algorithm}

In the workload fine-tuning stage, we need to optimize the partition index such that each device has a similar workload. We define the workload of $j$-th layer on the $i$-th device as \texttt{dt[i]} * \texttt{lf[j]} + \texttt{ct[i]}. The total workload on a device is then the sum of the workload of the layers allocated to it. 

The workload is used as a heuristic to guide each device to adjust its allocated layers. We adopted an iterative approach to optimize the load balance, by calculating the workload on all the devices and setting the average workload as the target workload. If a device has a workload that is less than the target, it should take one more layer from the next device. If the device has a workload that is higher than the target, it needs to consider whether the next device has a workload below the target value. If so, it is better to give away its last layer to the next device. 

To prevent the device workload from wandering around the target value, the device will only give away its last layer when its remaining workload is still higher than the target workload. In this way, the devices with longer benchmark time (weaker computation power) will be allocated with fewer layers and vice versa. While the layer distribution is changing, it is always necessary to make sure the memory requirement is still fulfilled so as not to step into the out-of-memory problem. When the maximum iteration is reached or the partition index does not change anymore, we can use the final partition index to allocate the layers to the devices. The full flow of the algorithm is shown in Algorithm \ref{algo:fine-allocation}.

\begin{algorithm}
\SetAlgoLined
\SetKwInOut{Data}{Data}
\SetKwInOut{Result}{Result}
\Data{~dt, dm, ct, lf, lm, pi, max\_iter}
\Result{~pi}
 
counter = 0\;
\texttt{\\}

\While{counter < max\_iter}{
    target = compute\_mean\_workload(dt, lf, ct, pi)\;
    \texttt{\\}
    
    \For{$i \gets 0$  \KwTo $D$-2} {
        
        \If{this\_device\_workload $<$ target \&\& get\_num\_layer(pi, $i$+1) $>$ 1 \&\& is\_memory\_satisfied\_after\_change}{
            pi[$i$+1] += 1\;
        }
        \Else{
            \If{ next\_device\_workload $<$ target \&\&
                this\_device\_workload $-$ this\_device\_last\_layer\_workload $>$ target \&\&
                get\_num\_layer(pi, $i$) $>$ 1 \&\&
                is\_memory\_satisfied\_after\_change
                }{
                pi[$i$+1] -= 1\;
            }
        }
        
    }
    
    \uIf{is\_pi\_unchanged $||$ counter == max\_iter}{
        return pi\;
    }
    
    counter += 1
}

\caption{Optimize workload}
\label{algo:fine-allocation}
\end{algorithm}

\section{Implementation} 

To integrate model parallelism, load balance optimization, and federated learning, we implemented a Python library called SCAELUM, which stands for SCAling Extreme Large models with Unified geo-distributed Mechanism. SCAELUM is built on PyTorch \cite{NEURIPS2019_9015} and provides a simple API to establish model parallelism for training using PyTorch remote procedure call (RPC).

\begin{figure}
    \centering
    \includegraphics[width=0.9\columnwidth]{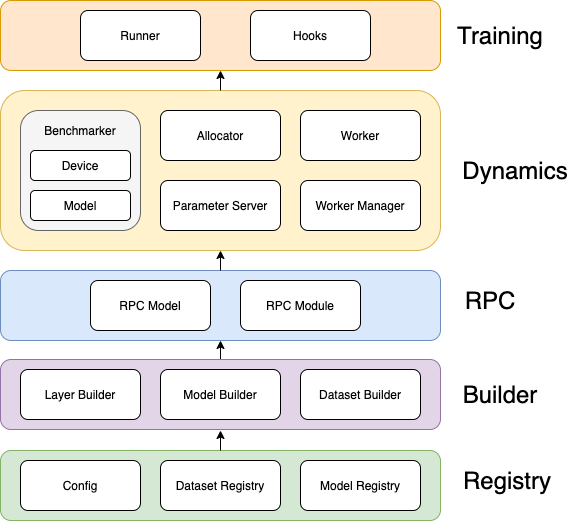}
    \caption{Architecture of SCAELUM library}
    \label{fig:arch-dllb}
\end{figure}

\subsection{Architecture of SCAELUM} 
The layered architecture of SCAELUM has shown in Figure \ref{fig:arch-dllb}. This architecture splits the library into individual modules and makes the library flexible and maintainable.

The Registry Layer contains modules for various datasets and deep learning models. The Dataset Registry currently contains three dataset modules for CIFAR10 dataset \cite{Krizhevsky2009LearningML}, GLUE dataset \cite{wang-etal-2018-glue} as well as random two-dimension floating tensors. In the Model Registry, we have registered the commonly-used layers in PyTorch as well as the various layers of ResNet \cite{7780459} and BERT\cite{devlin-etal-2019-bert}. If the user wants to add the customized module to this framework, he can register the customized module to the respective registry easily.

The Builder Layer relies on the Registry Layer to build the objects involved in training. Users can specify the architecture of the training model by declaring the layers in a sequential manner as well as the parameters for the data loader in the configuration file. The corresponding builder will take in the configuration, which is usually a Python dictionary, and create an object. In this way, users can conduct different experiments, only need to modify the configuration file.

The RPC Layer provides wrappers for the normal PyTorch modules for easy RPC calls. The wrapper module objects are instantiated remotely on the workers and provide RPC-related methods for the central server to call. For example, in RPC-based distributed training, PyTorch requires the central server to hold the references to all trainable parameters for back-propagation. The wrapper module can collect the allocated trainable parameters on the worker and pass their references back to the central server. Moreover, the wrapper module provides an interface to interact with the central server to distribute and gather weights to save the distributed weights into a single checkpoint file.

The Dynamics Layer is the core of the SCAELUM library and is designed to allocate the layers to the workers. The Parameter Server object hosts the full copy of the model weights and the Allocator is tasked to distribute the weights to the workers. The allocator supports three options, which are even allocation, heuristic-based Sky Computing, and optimal allocation calculated by Gurobi \cite{gurobi} solver. Since the allocation problem is NP-hard, we provide the optimal allocation option only for experimental purposes. When the last two strategies are selected, the Benchmarker will take charge to gather information about the devices and the training model for allocation.

Lastly, the Training Layer provides a Runner to conduct RPC-based distributed training, and Hook modules are provided to control the training workflow. For example, the CheckpointHook component allows the runner to save the distributed weights into a single file at a fixed interval.

\subsection{Simulating Heterogeneous Systems} 

As the devices involved in federated learning have different computing power, the whole system can be seen as a heterogeneous system. However, in most cases, the HPC cluster havs the same hardware specifications. To address this difference, we implemented speed control in the RPC module of SCAELUM to artificially adjust the computing power of the device. An iteration of training includes a forward pass and a backward pass. Our strategy is to measure the time taken in the forward and backward pass and let the process sleep for a period of time. By setting the value of the parameter \emph{slow\_down}, the sleep time is calculated by \emph{slow\_down} $\times$ \emph{forward\_time} or \emph{slow\_down} $\times$ \emph{backward\_time}. For example, if \emph{slow\_down} = 1, the time for forward and backward computation will be 2 times the original value.

It is easy to implement this in the forward pass as we can directly measure the time in the \emph{forward} method of the PyTorch module. Meanwhile, it is more difficult to measure the time in the backward pass since the backward computation is handled automatically by the autograd package. To tackle this problem, we applied our customized autograd function to the module. This autograd function comes with a distributed timer which measures the time taken to run the backward pass on this module. In this way, we can slow down both the forward and backward pass to simulate devices with variant computing power.

\begin{figure} 
    \centering
    \includegraphics[width=0.9\columnwidth]{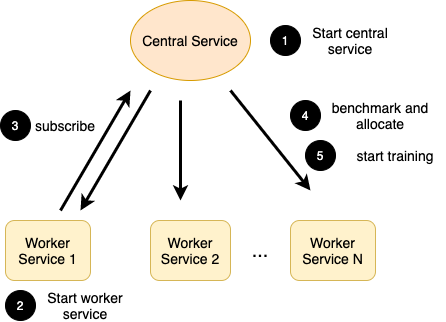}
    \caption{Workflow of DLLB-Fed}
    \label{fig:dllb-fed}
\end{figure}

\subsection{Training using SCAELUM}
The SCAELUM library provides the necessary modules for model parallelism training with load balance optimization. We provided two methods to utilize SCAELUM for experiments.

\subsubsection{Training with SCAELUM-Fed} \texttt{\\}

We designed and implemented a new testing framework called SCAELUM-Fed which uses SCAELUM to simulate the real federated learning scenario. SCAELUM-Fed wraps the central server and the user's device as individual services and provides a user-friendly command-line interface to interact with these services. The workflow of SCAELUM-Fed is shown in Figure \ref{fig:dllb-fed}. 

To use SCAELUM-Fed for training, the user first needs to start a central service. The central service contains the core modules of SCAELUM such as benchmarkers, allocator, worker manager, and runner. The worker manager of the central service will keep track of the worker service participating in training. Next, one or more worker services can be started. These worker services should subscribe to the central service, registering themselves to the worker manager of the central service. Once the central service has received subscriptions from some workers, it can start to gather information about the training model and the worker services by conducting benchmarking tests. The central service will then allocate the model using the Sky Computing. At last, the central service can send the request to all the subscribed worker services to start model-parallel distributed training. All the actions can be done via the command-line interface. By using SCAELUM-Fed, we can simulate how users' devices interact with the central server and conduct experiments to evaluate the effectiveness of our load balance optimization algorithm by adding or removing the worker service.

\subsubsection{Training with MPI} \texttt{\\}

One disadvantage of SCAELUM-Fed is that it becomes troublesome to manage the worker services by command line when the number of worker services becomes large. It is reasonably not a good choice if we wish to explore the performance of our allocation framework on large-scale distributed systems. To address this issue, instead of running some services, we extract the workflow from SCAELUM-Fed and use MPI to launch multiple processes on supercomputers.

\section{Experiments}
\subsection{Hardware Specification} 
To evaluate the effectiveness of the Sky Computing, we conducted experiments on an HPC cluster where each node has one P100 GPU. We chose to run our experiments on 16, 32, and 64 compute nodes. As the central parameter server needs to take one GPU, the number of devices that participate in federated learning is $D-1$ where $D$ is the total number of GPUs used. Even though this does not make the number of devices a multiple of two, our experiments still demonstrate the effectiveness of our algorithm.

\subsection{Task Description} 
\begin{figure} 
    \centering
    \includegraphics[width=0.9\columnwidth]{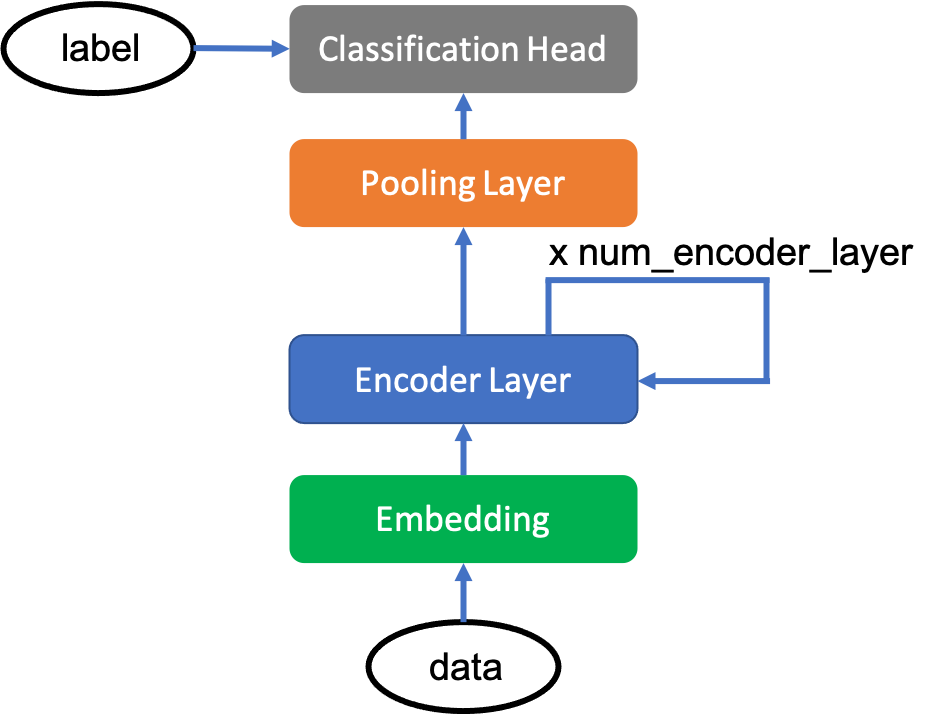}
    \caption{Architecture of BERT}
    \label{fig:arch-bert}
\end{figure}

In our experiments, we chose BERT \cite{devlin-etal-2019-bert} as the model to tackle the MNLI task \cite{N18-1101} on GLUE dataset \cite{wang-etal-2018-glue}. MNLI stands for Multi-Genre Natural Language Inference. In this task, we give a pair of sentences as input data to BERT and classify whether the second sentence is a contradiction, entailment, or neutral statement of the first premise sentence. The architecture of BERT can be split into the embedding layer, the encoder layers, the pooling layer, and the classification head as shown in Figure \ref{fig:arch-bert}. The encoder layer can be further divided into the self-attention layer, the intermediate layer, and the output layer as discussed in Figure \ref{fig:arch-bert-encoder-layer} and it can be repeated infinitely since the input and output have the same shape. Therefore, we can change the number of encoder layers in BERT to have a different amount of computation when we change the scale of our experiments. In the official BERT-Large model, 24 encoder layers are used and each layer has 16 attention heads. In our experiments, we use a maximum of 160 encoder layers, which is almost 7 times larger than the BERT-Large model. We stick to 16 attention heads as used in the official configurations and train the model with a batch size of 32.

\subsection{Memory Control}
When allocating the layers to devices, the essential condition is that the memory usage does not exceed the memory limit on the device to avoid the out-of-memory problem. This is critical as training will fail with insufficient memory. Our framework ensures the memory limit is not exceeded and the training can be executed successfully.

\subsection{Speed Control} 
To simulate devices with different computational power, we multiply the forward time and backward time with a reproducible random-generated penalty factor \emph{slow\_down}. In our experiments, the values of \emph{slow\_down} are between 1 and 7. The larger the penalty factor, the slower the device. The adjusted computational power of devices is normally distributed. This allows us to observe the performance of our algorithm in a heterogeneous-like setting.

To measure the computation power of the devices in federated learning, the central server sends a request to each device and runs a simple benchmark test. In our experiments, we run the forward pass of a 10-layer convolutional neural network for 30 iterations. The input data has a shape of (32, 256, 64, 64) while the convolution layer has the same input and output dimension of 256 with a kernel size of 3 and padding size of 1.

\subsection{Communication Control} 
In model parallelism, P2P communication is used when passing tensors between devices, and the communication latency, which depends on the physical distance between two devices, cannot be ignored. Therefore, we test the communication time at the benchmark stage and treat it as a part of the performance of the devices. In the experiments, all experiments were run in the same SLURM job on the same HPC cluster and this ensures the same latency between consecutive workers for the sake of fairness.

\subsection{Overall Results}
\begin{figure}
    \centering
    \includegraphics[width=\columnwidth]{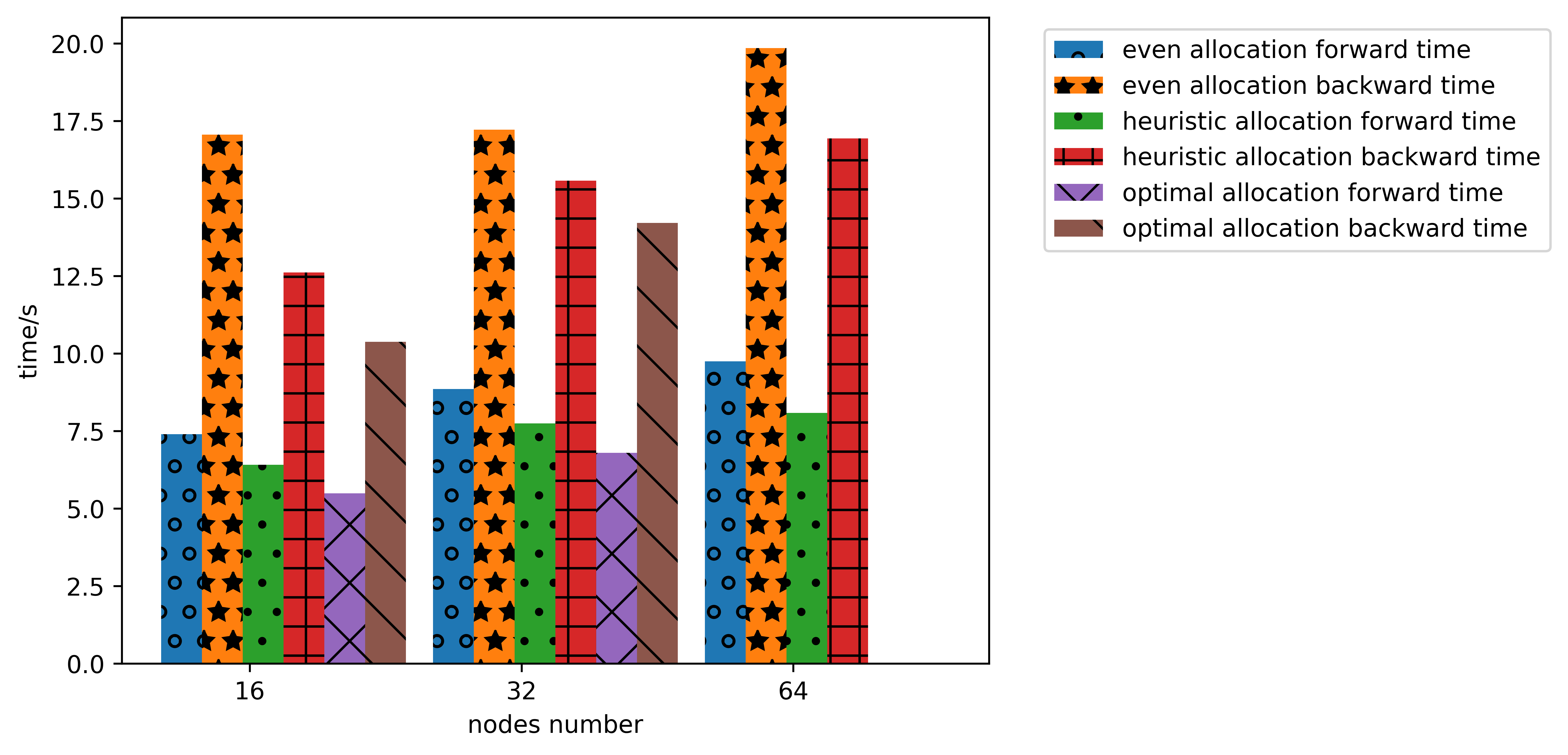}
    \caption{Average Forward/Backward Time}
    \label{fig:train_time_per_iter}
\end{figure}

From the training results in Figure \ref{fig:train_time_per_iter}, it can be observed that the Sky Computing outperforms the even allocation strategy in all scales. Because the allocation problem is NP-hard, it will consume an unnecessary huge amount of power and take an unbearable long time to wait when scale grows large. So there is no need to figure out an optimal solution by using significant power, thus we only apply optimal allocation up to 32 nodes.

\subsection{Strong Scaling}
 In strong scaling experiments, we used a very large BERT model by setting the number of encoder layers to be 80 so that we have 403 discrete layers in total. By keeping the model size constant, we use 16, 32, and 64 nodes for training. Since model parallelism by layer is adopted, the communication time for an iteration is expected to increase with an increasing number of devices participating in training.

\begin{figure}
    \centering
    \includegraphics[width=\columnwidth]{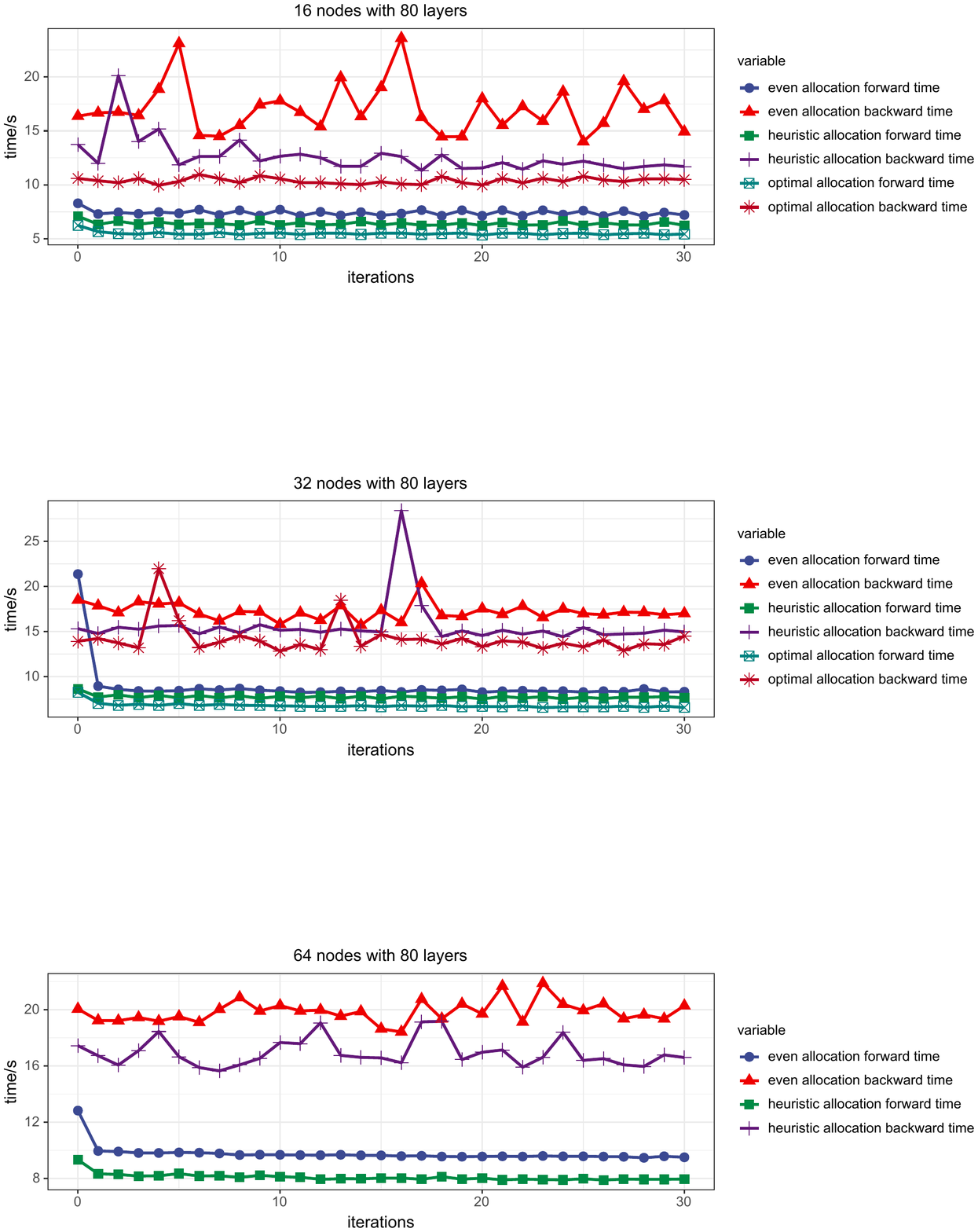}
    \caption{Strong Scaling Results}
    \label{fig:strong-scaling}
\end{figure}


Figure \ref{fig:strong-scaling} shows the forward time and backward time for 30 iterations during training. The forward and backward times are lower with the Sky Computing for all cases. The optimal allocation can reduce 35\%, 19.4\% training time for 16, 32 nodes respectively. And the heuristic allocation can reduce 22.2\%, 10.5\%, 15.4\% training time for 16, 32, 64 nodes respectively.

\subsection{Weak Scaling} 
To perform weak scaling experiments, we increase the number of encoder layers and the number of devices at the same rate to keep the problem size on each device constant. Even though the embedding layer, pooling layer, and the classification head cannot be repeated proportionally, the increase in the total number of layers is still approximately linear. 

\begin{figure}
    \centering
    \includegraphics[width=\columnwidth]{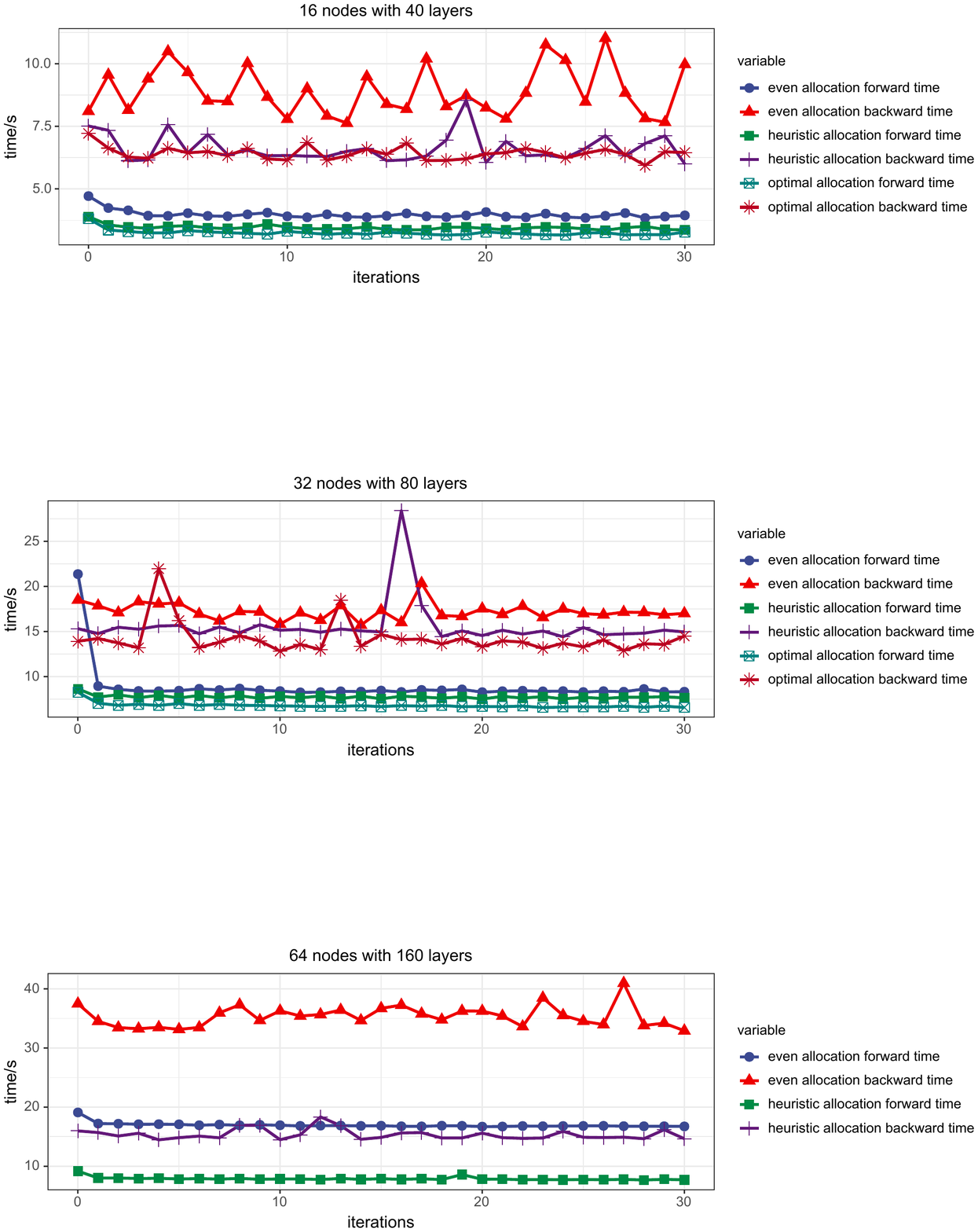}
    \caption{Weak Scaling Results}
    \label{fig:weak-scaling}
\end{figure}

We used the same number of nodes as the strong scaling experiments and set the number of encoder layers to be 40, 80, and 160. The run time results for the weak scaling experiments are reported in Figure \ref{fig:weak-scaling}. In all experiments, our load-balanced allocation strategy can outperform the even allocation counterpart by a large margin. The optimal allocation can reduce 25\%, 19.4\% training time for 16, 32 nodes respectively. And the heuristic allocation can reduce 21.7\%, 10.5\%, 55.6\% training time for 16, 32, 64 nodes respectively.

\section{Discussion}

\subsection{Real-world Geo-distributed Computing}

As a new type of heterogeneous computing, geo-distributed computing has attracted more and more attention. In our paper, we use HPC to simulate a real-world situation by setting normally distributed factors. However, we also noticed some flaws caused by this stimulation:

\begin{enumerate}
    \item Communication Latency: Since different heterogeneous devices are often connected through the Internet, communication latency is extremely high and it will dominate the overall time cost in some cases. In our simulation, the devices are connected by a high-speed local area network, which currently is not consistent with the actual situation.
    \item Computing Power: In real-world scenarios, the computing power of devices may vary widely, especially with the continuous development of SoCs in the past years, the graphics performance of smart devices has been explosively improved. According to GeekBench\footnote{https://browser.geekbench.com/ios-benchmarks}, iPhone's graphics performance quadrupled in four years and will continue to grow for the foreseeable future. As a result, the heterogeneity of computing speed will be more significant in real scenarios, and we only do a preliminary simulation in our experiments
    \item Available Memory: Due to the high homogeneity of HPC hardware, the difference in available memory obtained by the benchmark stage is almost imperceptible, but in the real world, there may be some specific differences.
\end{enumerate}

Sky Computing is currently under heavy development, and many of its features have not been added. In the next, we will first focus on fixing these problems and providing a user-friendly system. With this project, we hope to extend Sky Computing to more fields and bring some new inspiration to geo-distributed computing. 



\subsection{Data privacy}
One difference between our training set and the general federated learning is that the feature map is passed between workers to perform the feed-forward computation. One concern raised is whether data privacy is still preserved in our case. Three reasons are provided below to explain why user privacy is not infringed even though there is data passing between the devices.
\begin{itemize}
  \item In traditional federated learning, each device will hold a full copy of the model while in our case, each device only holds a part of the model. When a worker receives a feature map from the previous worker, it does not have information about the model architecture and weights on the previous worker. Thus, it cannot compute reversely to recover the input data generated by the user.
  \item To make sure the data is securely transferred, the intermediate feature map must be encrypted when the device sends out the feature map to the next device. Before sending out the feature map, other existing techniques such as differential privacy \cite{10.1145/3438872.3439097} can be applied to mask the feature map as well. In this way, it becomes even more difficult to recover useful information from the transferred data.
  \item In our current framework, only one device is receiving the raw input data as it holds the first layer of the model. Meanwhile, all devices have the raw user data in reality. However, we should not transfer the raw user data from one device to another device to prevent possible data leakage. To solve this problem, we propose that each device should at least hold a copy of the first layer. In this way, the device only passes the output of the first layer to other workers when raw user data is used for training. The output feature map of the first layer can be passed securely to other workers with encryption or differential privacy. The raw user data is still kept intact on the user's device to safeguard their privacy.
\end{itemize}


\section{Conclusion}
In this paper, we proposed a method to conduct training of large-scale models in federating learning and proposed a method that optimizes the load balance in order to remove the computation bottleneck and increase training speed. Our Sky Computing guides the devices to decide the layers they should fetch from the parameter server. In heuristic allocation, it iteratively updates the model allocation to reduce the average workload on devices based on both the model and device information. As shown in our experiments, we compared the performance produced by the naive even-allocation method and our Sky Computing. Our framework provides a consistent speedup for the training and outperforms the naive approach by 55.0\% when training 160-layer BERT on 64 nodes.

\bibliographystyle{IEEEtran}
\bibliography{reference.bib}

\begin{thebibliography}{10}
\providecommand{\url}[1]{#1}
\csname url@samestyle\endcsname
\providecommand{\newblock}{\relax}
\providecommand{\bibinfo}[2]{#2}
\providecommand{\BIBentrySTDinterwordspacing}{\spaceskip=0pt\relax}
\providecommand{\BIBentryALTinterwordstretchfactor}{4}
\providecommand{\BIBentryALTinterwordspacing}{\spaceskip=\fontdimen2\font plus
\BIBentryALTinterwordstretchfactor\fontdimen3\font minus
  \fontdimen4\font\relax}
\providecommand{\BIBforeignlanguage}[2]{{%
\expandafter\ifx\csname l@#1\endcsname\relax
\typeout{** WARNING: IEEEtran.bst: No hyphenation pattern has been}%
\typeout{** loaded for the language `#1'. Using the pattern for}%
\typeout{** the default language instead.}%
\else
\language=\csname l@#1\endcsname
\fi
#2}}
\providecommand{\BIBdecl}{\relax}
\BIBdecl

\bibitem{devlin-etal-2019-bert}
\BIBentryALTinterwordspacing
J.~Devlin, M.-W. Chang, K.~Lee, and K.~Toutanova, ``{BERT}: Pre-training of
  deep bidirectional transformers for language understanding,'' in
  \emph{Proceedings of the 2019 Conference of the North {A}merican Chapter of
  the Association for Computational Linguistics: Human Language Technologies,
  Volume 1 (Long and Short Papers)}.\hskip 1em plus 0.5em minus 0.4em\relax
  Minneapolis, Minnesota: Association for Computational Linguistics, Jun. 2019,
  pp. 4171--4186. [Online]. Available:
  \url{https://www.aclweb.org/anthology/N19-1423}
\BIBentrySTDinterwordspacing

\bibitem{NEURIPS2020_1457c0d6}
\BIBentryALTinterwordspacing
T.~Brown, B.~Mann, N.~Ryder, M.~Subbiah, J.~D. Kaplan, P.~Dhariwal,
  A.~Neelakantan, P.~Shyam, G.~Sastry, A.~Askell, S.~Agarwal, A.~Herbert-Voss,
  G.~Krueger, T.~Henighan, R.~Child, A.~Ramesh, D.~Ziegler, J.~Wu, C.~Winter,
  C.~Hesse, M.~Chen, E.~Sigler, M.~Litwin, S.~Gray, B.~Chess, J.~Clark,
  C.~Berner, S.~McCandlish, A.~Radford, I.~Sutskever, and D.~Amodei, ``Language
  models are few-shot learners,'' in \emph{Advances in Neural Information
  Processing Systems}, H.~Larochelle, M.~Ranzato, R.~Hadsell, M.~F. Balcan, and
  H.~Lin, Eds., vol.~33.\hskip 1em plus 0.5em minus 0.4em\relax Curran
  Associates, Inc., 2020, pp. 1877--1901. [Online]. Available:
  \url{https://proceedings.neurips.cc/paper/2020/file/1457c0d6bfcb4967418bfb8ac142f64a-Paper.pdf}
\BIBentrySTDinterwordspacing

\bibitem{44873}
\BIBentryALTinterwordspacing
G.~Hinton, O.~Vinyals, and J.~Dean, ``Distilling the knowledge in a neural
  network,'' in \emph{NIPS Deep Learning and Representation Learning Workshop},
  2015. [Online]. Available: \url{http://arxiv.org/abs/1503.02531}
\BIBentrySTDinterwordspacing

\bibitem{8953493}
H.~{Kim}, M.~U.~K. {Khan}, and C.~{Kyung}, ``Efficient neural network
  compression,'' in \emph{2019 IEEE/CVF Conference on Computer Vision and
  Pattern Recognition (CVPR)}, 2019, pp. 12\,561--12\,569.

\bibitem{McMahan2017CommunicationEfficientLO}
H.~B. McMahan, E.~Moore, D.~Ramage, S.~Hampson, and B.~A.~Y. Arcas,
  ``Communication-efficient learning of deep networks from decentralized
  data,'' in \emph{AISTATS}, 2017.

\bibitem{dong2021attention}
Y.~Dong, J.-B. Cordonnier, and A.~Loukas, ``Attention is not all you need: Pure
  attention loses rank doubly exponentially with depth,'' 2021.

\bibitem{10.1145/3320060}
\BIBentryALTinterwordspacing
T.~Ben-Nun and T.~Hoefler, ``Demystifying parallel and distributed deep
  learning: An in-depth concurrency analysis,'' \emph{ACM Comput. Surv.},
  vol.~52, no.~4, Aug. 2019. [Online]. Available:
  \url{https://doi.org/10.1145/3320060}
\BIBentrySTDinterwordspacing

\bibitem{NEURIPS2019_093f65e0}
\BIBentryALTinterwordspacing
Y.~Huang, Y.~Cheng, A.~Bapna, O.~Firat, D.~Chen, M.~Chen, H.~Lee, J.~Ngiam,
  Q.~V. Le, Y.~Wu, and z.~Chen, ``Gpipe: Efficient training of giant neural
  networks using pipeline parallelism,'' in \emph{Advances in Neural
  Information Processing Systems}, H.~Wallach, H.~Larochelle, A.~Beygelzimer,
  F.~d\textquotesingle Alch\'{e}-Buc, E.~Fox, and R.~Garnett, Eds.,
  vol.~32.\hskip 1em plus 0.5em minus 0.4em\relax Curran Associates, Inc.,
  2019. [Online]. Available:
  \url{https://proceedings.neurips.cc/paper/2019/file/093f65e080a295f8076b1c5722a46aa2-Paper.pdf}
\BIBentrySTDinterwordspacing

\bibitem{10.1145/3341301.3359646}
\BIBentryALTinterwordspacing
D.~Narayanan, A.~Harlap, A.~Phanishayee, V.~Seshadri, N.~R. Devanur, G.~R.
  Ganger, P.~B. Gibbons, and M.~Zaharia, ``Pipedream: Generalized pipeline
  parallelism for dnn training,'' in \emph{Proceedings of the 27th ACM
  Symposium on Operating Systems Principles}, ser. SOSP '19.\hskip 1em plus
  0.5em minus 0.4em\relax New York, NY, USA: Association for Computing
  Machinery, 2019, p. 1–15. [Online]. Available:
  \url{https://doi.org/10.1145/3341301.3359646}
\BIBentrySTDinterwordspacing

\bibitem{shoeybi2020megatronlm}
M.~Shoeybi, M.~Patwary, R.~Puri, P.~LeGresley, J.~Casper, and B.~Catanzaro,
  ``Megatron-lm: Training multi-billion parameter language models using model
  parallelism,'' 2020.

\bibitem{NEURIPS2019_9d63484a}
\BIBentryALTinterwordspacing
S.~Yun, M.~Jeong, R.~Kim, J.~Kang, and H.~J. Kim, ``Graph transformer
  networks,'' in \emph{Advances in Neural Information Processing Systems},
  H.~Wallach, H.~Larochelle, A.~Beygelzimer, F.~d\textquotesingle
  Alch\'{e}-Buc, E.~Fox, and R.~Garnett, Eds., vol.~32.\hskip 1em plus 0.5em
  minus 0.4em\relax Curran Associates, Inc., 2019. [Online]. Available:
  \url{https://proceedings.neurips.cc/paper/2019/file/9d63484abb477c97640154d40595a3bb-Paper.pdf}
\BIBentrySTDinterwordspacing

\bibitem{47976}
\BIBentryALTinterwordspacing
K.~A. Bonawitz, H.~Eichner, W.~Grieskamp, D.~Huba, A.~Ingerman, V.~Ivanov,
  C.~M. Kiddon, J.~Konečný, S.~Mazzocchi, B.~McMahan, T.~V. Overveldt,
  D.~Petrou, D.~Ramage, and J.~Roselander, ``Towards federated learning at
  scale: System design,'' in \emph{SysML 2019}, 2019, to appear. [Online].
  Available: \url{https://arxiv.org/abs/1902.01046}
\BIBentrySTDinterwordspacing

\bibitem{10.1145/3298981}
\BIBentryALTinterwordspacing
Q.~Yang, Y.~Liu, T.~Chen, and Y.~Tong, ``Federated machine learning: Concept
  and applications,'' \emph{ACM Trans. Intell. Syst. Technol.}, vol.~10, no.~2,
  Jan. 2019. [Online]. Available: \url{https://doi.org/10.1145/3298981}
\BIBentrySTDinterwordspacing

\bibitem{NIPS2014_1ff1de77}
\BIBentryALTinterwordspacing
M.~Li, D.~G. Andersen, A.~J. Smola, and K.~Yu, ``Communication efficient
  distributed machine learning with the parameter server,'' in \emph{Advances
  in Neural Information Processing Systems}, Z.~Ghahramani, M.~Welling,
  C.~Cortes, N.~Lawrence, and K.~Q. Weinberger, Eds., vol.~27.\hskip 1em plus
  0.5em minus 0.4em\relax Curran Associates, Inc., 2014. [Online]. Available:
  \url{https://proceedings.neurips.cc/paper/2014/file/1ff1de774005f8da13f42943881c655f-Paper.pdf}
\BIBentrySTDinterwordspacing

\bibitem{NEURIPS2019_9015}
\BIBentryALTinterwordspacing
A.~Paszke, S.~Gross, F.~Massa, A.~Lerer, J.~Bradbury, G.~Chanan, T.~Killeen,
  Z.~Lin, N.~Gimelshein, L.~Antiga, A.~Desmaison, A.~Kopf, E.~Yang, Z.~DeVito,
  M.~Raison, A.~Tejani, S.~Chilamkurthy, B.~Steiner, L.~Fang, J.~Bai, and
  S.~Chintala, ``Pytorch: An imperative style, high-performance deep learning
  library,'' in \emph{Advances in Neural Information Processing Systems 32},
  H.~Wallach, H.~Larochelle, A.~Beygelzimer, F.~d\textquotesingle
  Alch\'{e}-Buc, E.~Fox, and R.~Garnett, Eds.\hskip 1em plus 0.5em minus
  0.4em\relax Curran Associates, Inc., 2019, pp. 8024--8035. [Online].
  Available:
  \url{http://papers.neurips.cc/paper/9015-pytorch-an-imperative-style-high-performance-deep-learning-library.pdf}
\BIBentrySTDinterwordspacing

\bibitem{Krizhevsky2009LearningML}
A.~Krizhevsky, ``Learning multiple layers of features from tiny images,'' 2009.

\bibitem{wang-etal-2018-glue}
\BIBentryALTinterwordspacing
A.~Wang, A.~Singh, J.~Michael, F.~Hill, O.~Levy, and S.~Bowman, ``{GLUE}: A
  multi-task benchmark and analysis platform for natural language
  understanding,'' in \emph{Proceedings of the 2018 {EMNLP} Workshop
  {B}lackbox{NLP}: Analyzing and Interpreting Neural Networks for {NLP}}.\hskip
  1em plus 0.5em minus 0.4em\relax Brussels, Belgium: Association for
  Computational Linguistics, Nov. 2018, pp. 353--355. [Online]. Available:
  \url{https://www.aclweb.org/anthology/W18-5446}
\BIBentrySTDinterwordspacing

\bibitem{7780459}
K.~{He}, X.~{Zhang}, S.~{Ren}, and J.~{Sun}, ``Deep residual learning for image
  recognition,'' in \emph{2016 IEEE Conference on Computer Vision and Pattern
  Recognition (CVPR)}, 2016, pp. 770--778.

\bibitem{gurobi}
\BIBentryALTinterwordspacing
{Gurobi Optimization, LLC}, ``{Gurobi Optimizer Reference Manual},'' 2021.
  [Online]. Available: \url{https://www.gurobi.com}
\BIBentrySTDinterwordspacing

\bibitem{N18-1101}
\BIBentryALTinterwordspacing
A.~Williams, N.~Nangia, and S.~Bowman, ``A broad-coverage challenge corpus for
  sentence understanding through inference,'' in \emph{Proceedings of the 2018
  Conference of the North American Chapter of the Association for Computational
  Linguistics: Human Language Technologies, Volume 1 (Long Papers)}.\hskip 1em
  plus 0.5em minus 0.4em\relax Association for Computational Linguistics, 2018,
  pp. 1112--1122. [Online]. Available:
  \url{http://aclweb.org/anthology/N18-1101}
\BIBentrySTDinterwordspacing

\bibitem{10.1145/3438872.3439097}
\BIBentryALTinterwordspacing
Z.~Chuanxin, S.~Yi, and W.~Degang, ``Federated learning with gaussian
  differential privacy,'' in \emph{Proceedings of the 2020 2nd International
  Conference on Robotics, Intelligent Control and Artificial Intelligence},
  ser. RICAI 2020.\hskip 1em plus 0.5em minus 0.4em\relax New York, NY, USA:
  Association for Computing Machinery, 2020, p. 296–301. [Online]. Available:
  \url{https://doi.org/10.1145/3438872.3439097}
\BIBentrySTDinterwordspacing

\end{thebibliography}

\end{document}